\documentclass[11pt,a4paper]{article}
\usepackage[nohyperref]{emnlp2018}
\usepackage{times}
\usepackage{latexsym}
\usepackage[utf8]{inputenc}
\usepackage{natbib}
\usepackage{graphicx}

\aclfinalcopy 

\graphicspath{ {./} }

\usepackage{booktabs}
\usepackage[hyphens]{url}
\usepackage{amssymb}

\usepackage{amsmath}
\usepackage{bm}
\usepackage{mathtools}
\usepackage{paralist}
\usepackage[labelformat=simple]{subfig}
\usepackage{booktabs}
\usepackage{multirow}
\usepackage{microtype}
\usepackage{tabularx}
\usepackage{hyperref}
\usepackage{xspace}
\usepackage{xcolor}
\usepackage{tablefootnote}
\usepackage{subfig}
\usepackage{capt-of}

\newcommand\MyHead[2]{%
  \multicolumn{1}{l}{\parbox{#1}{\centering #2}}
}

\newcommand{\lstm}{\textit{LSTM}\xspace}
\newcommand{\bilstm}{\textit{BiLSTM}\xspace}
\newcommand{\bow}{\textit{Bow}\xspace}
\newcommand{\cnn}{\textit{CNN}\xspace}
\newcommand{\averaging}{\textit{Ave}\xspace}

\newcommand{\SanOne}{{\textit{\begin{tabular}{@{}c@{}}1-Layer \\ SSAN + RPR\end{tabular}}}}
\newcommand{\SanTwo}{{\textit{\begin{tabular}{@{}c@{}}2-Layer \\ SSAN + RPR\end{tabular}}}}
\newcommand{\SanOneNoPos}{{\textit{\begin{tabular}{@{}c@{}}1-Layer \\ SSAN\end{tabular}}}}
\newcommand{\SanOneWithPosEnc}{{\textit{\begin{tabular}{@{}c@{}}1-Layer \\ SSAN + PE\end{tabular}}}}
\newcommand{\Transf}{{\textit{\begin{tabular}{@{}c@{}}Transformer\\ Encoder + RPR\end{tabular}}}}
\newcommand{\TransfWithPosEnc}{{\textit{\begin{tabular}{@{}c@{}}Transformer\\ Encoder + PE\end{tabular}}}}

\newcommand{\kq}{q}

\newcommand{\dmodel}{d_{\text{model}}}
\newcommand{\dclasses}{d_{\text{classes}}}

\newcommand{\rt}[1]{\rotatebox{90}{#1}}
\newcommand{\rrt}[1]{\rotatebox{45}{#1}}

\newcommand{\sd}[1]{\par \tiny #1}
\usepackage{scalefnt}

\title{Self-Attention: A Better Building Block for Sentiment Analysis Neural Network Classifiers}

\author{Artaches Ambartsoumian \\
  School of Computing Science \\
  Simon Fraser University \\
  Burnaby, BC, CANADA \\
  {\tt aambarts@sfu.ca} \\\And
  Fred Popowich \\
  School of Computing Science \\
  Simon Fraser University \\
  Burnaby, BC, CANADA \\
  {\tt popowich@sfu.ca} \\}
\date{}

\begin{document}
\maketitle


\begin{abstract}
  
  
  Sentiment Analysis has seen much progress in the past two decades. For the past few years, neural network approaches, primarily RNNs and CNNs, have been the most successful for this task. Recently, a new category of neural networks, self-attention networks (SANs), have been created which utilizes the attention mechanism as the basic building block. Self-attention networks have been shown to be effective for sequence modeling tasks, while having no recurrence or convolutions. In this work we explore the effectiveness of the SANs for sentiment analysis. We demonstrate that SANs are superior in performance to their RNN and CNN counterparts by comparing their classification accuracy on six datasets as well as their model characteristics such as training speed and memory consumption. Finally, we explore the effects of various SAN modifications such as multi-head attention as well as two methods of incorporating sequence position information into SANs.

  
  
\end{abstract}


\section{Introduction}

Sentiment analysis, also know as opinion mining, deals with determining the opinion classification of a piece of text. Most commonly the classification is whether the writer of a piece of text is expressing a position or negative attitude towards a product or a topic of interest.  Having more than two sentiment classes is called fine-grained sentiment analysis with the extra classes representing intensities of positive/negative sentiment (e.g. very-positive) and/or the neutral class. This field has seen much growth for the past two decades, with many applications and multiple classifiers proposed \cite{MANTYLA201816}. Sentiment analysis has been applied in areas such as social media \cite{jansen2009twitter}, movie reviews \cite{Pang:2002:TUS:1118693.1118704}, commerce \cite{jansen2009twitter}, and health care \cite{greaves2013use} \cite{greaves2013harnessing}.

In the past few years, neural network approaches have consistently advanced the state-of-the-art technologies for sentiment analysis and other natural language processing (NLP) tasks. For sentiment analysis, the neural network approaches typically use pre-trained word embeddings such as word2vec \cite{mikolov2013efficient} or GloVe\cite{pennington2014glove} for input, which get processed by the model to create a sentence representation that is finally used for a softmax classification output layer. The main neural network architectures that have been applied for sentiment analysis are recurrent neural networks(RNNs) \cite{tai2015improved} and convolutional neural networks (CNNs) \cite{kim2014convolutional}, with RNNs being more popular of the two. For RNNs, typically gated cell variants such as long short-term memory (LSTM) \cite{hochreiter1997long},  Bi-Directional LSTM (BiLSTM) \cite{schuster1997bidirectional}, or gated recurrent unit (GRU) \cite{W14-4012_GRU} are used.

Most recently, Vaswani et al. \cite{Vaswani2017AttentionIA} introduced the first fully-attentional architecture, called Transformer, which utilizes only the self-attention mechanism and demonstrated its effectiveness on neural machine translation (NMT). The Transformer model achieved state-of-the-art performance on multiple machine translation datasets, without having recurrence or convolution components. Since then, self-attention networks have been successfully applied to a variety of tasks, including: image classification \cite{parmar2018image}, generative adversarial networks \cite{zhang2018self}, automatic speech recognition \cite{Povey2017ATS}, text summarization \cite{liu2018generating}, semantic role labeling \cite{Strubell2018LinguisticallyInformedSF}, as well as natural language inference and sentiment analysis \cite{shen2018disan}.

In this paper we demonstrate that self-attention is a better building block compared to recurrence or convolutions for sentiment analysis classifiers. We extend the work of \cite{SOTA_SA_2017} by exploring the behaviour of various self-attention architectures on six different datasets and making direct comparisons to their work. We set our baselines to be their results for \lstm, \bilstm, and \cnn models, and used the same code for dataset pre-processing, word embedding imports, and batch construction. Finally, we explore the effectiveness of SAN architecture variations such as different techniques of incorporating positional information into the network, using multi-head attention, and stacking self-attention layers. Our results suggest that relative position representations is superior to positional encodings, as well as highlight the efficiency of the stacking self-attention layers.

Source code is publicly available\footnote{\url{https://github.com/Artaches/SSAN-self-attention-sentiment-analysis-classification}}.


\section{Background}

The attention mechanism was introduced by \cite{bahdanau2014neural} to improve the RNN encoder-decoder sequence-to-sequence architecture for NMT \cite{NIPS2014_5346_seq2seq}. Since then, it has been extensively used to improve various RNN and CNN architectures (\cite{DBLP:journals/corr/ChengDL16}; \cite{kokkinos2017structural}; \cite{NIPS2016_6202}). The attention mechanism has been an especially popular modification for RNN-based architectures due to its ability to improve the modeling of long range dependencies (\cite{daniluk2017frustratingly}; \cite{AAAI1817176}).


\subsection{Attention}

Originally \cite{bahdanau2014neural} described attention as the process of computing a context vector for the next decoder step that contains the most relevant information from all of the encoder hidden states by performing a weighted average on the encoder hidden states. How much each encoder state contributes to the weighted average is determined by an alignment score between that encoder state and previous hidden state of the decoder. 

More generally, we can consider the previous decoder state as the query vector, and the encoder hidden states as key and value vectors. The output is a weighted average of the value vectors, where the weights are determined by the compatibility function between the query and the keys. Note that the keys and values can be different sets of vectors \cite{Vaswani2017AttentionIA}.

The above can be summarized by the following equations. Given a query $\kq$, values $(v_1, ..., v_n)$, and keys $(k_1, ..., k_n)$ we compute output $z$:

\begin{equation}\label{eq:attn}
z = \sum_{j=1}^{n} \alpha_{j} (v_j)
\end{equation}

\begin{equation}\label{eq:alpha}
\alpha_{j} = \frac{ \exp{f({k_j, \kq})} }{ \sum_{i=1}^{n} \exp{f({k_i, \kq})}}
\end{equation}
$\alpha_{j}$ is computed using the softmax function where $f(k_i, q)$ is the compatibility score between  $k_i$ and $\kq$,

For the compatibility function, we will be using using the scaled dot-product function from \cite{Vaswani2017AttentionIA}:

\begin{equation}\label{eq:e}
f(k, q) = \frac{(k)(q)^T}{\sqrt{d_k}}
\end{equation}
where $d_k$ is the dimension of the key vectors. This scaling is done to improve numerical stability as the dimension of keys, values, and queries grows.


\subsection{Self-Attention}


Self-attention is the process of applying the attention mechanism outlined above to every position of the source sequence. This is done by creating three vectors (query, key, value) for each sequence position, and then applying the attention mechanism for each position $x_i$, using the $x_i$ query vector and key and value vectors for all other positions. As a result, an input sequence $X=(x_1,x_2,...,x_n)$ of words is transformed into a sequence $Y=(y_1,y_2,...,y_n)$ where $y_i$ incorporates the information of $x_i$ as well as how $x_i$ relates to all other positions in $X$. The (query, key, value) vectors can be created by applying learned linear projections \cite{Vaswani2017AttentionIA}, or using feed-forward layers.

This computation can be done for the entire source sequence in parallel by grouping the queries, keys, and values in Q, K, V matrices\cite{Vaswani2017AttentionIA}.
\begin{equation}
   \mathrm{Attention}(Q, K, V) = \mathrm{softmax}(\frac{QK^T}{\sqrt{d_k}})V
\end{equation}


Furthermore, instead of performing self-attention once for (Q,K,V) of dimension $\dmodel$, \cite{Vaswani2017AttentionIA} proposed multi-head attention, which performs attention $h$ times on projected (Q,K,V) matrices of dimension $\dmodel/h$. For each head, the (Q,K,V) matrices are uniquely projected to dimension $\dmodel/h$ and self-attetnion is performed to yield an output of dimension $\dmodel/h$. The outputs of each head are then concatenated, and once again a linear projection layer is applied, resulting in an output of same dimensionality as performing self-attention once on the original (Q,K,V) matrices. This process is described by the following formulas:









\begin{align} 
   \mathrm{MultiHead}(Q, K, V) = \nonumber \hspace{25mm} \\
         \mathrm{Concat}(\mathrm{head_1}, ..., \mathrm{head_h})W^O
\end{align}
\begin{align} 
   \text{where}~\mathrm{head_i} = \nonumber \hspace{35mm} \\ 
        \mathrm{Attention}(QW^Q_i, KW^K_i, VW^V_i)
\end{align}




Where the projections are parameter matrices $W^Q_i \in \mathbb{R}^{\dmodel \times d_k}$, $W^K_i \in \mathbb{R}^{\dmodel \times d_k}$, $W^V_i \in \mathbb{R}^{\dmodel \times d_v}$ and $W^O \in \mathbb{R}^{h d_v \times \dmodel}$.

\subsection{Position Information Techniques}
The attention mechanism is completely invariant to sequence ordering, thus self-attention networks need to incorporate positional information. Three main techniques have been proposed to solve this problem: adding sinusoidal positional encodings or learned positional encoding to input embeddings, or using relative positional representations in the self-attention mechanism.

\subsubsection{Sinusoidal Position Encoding}
This method was proposed by \cite{Vaswani2017AttentionIA} to be used for the Transformer model. Here, positional encoding ($PE$) vectors are created using sine and cosine functions of difference frequencies and then are added to the input embeddings. Thus, the input embeddings and positional encodings must have the same dimensionality of ${\dmodel}$. The following sine and cosine functions are used:
\begin{align*}
    PE_{(pos,2i)} = sin(pos / 10000^{2i / \dmodel}) \\
    PE_{(pos,2i+1)} = cos(pos / 10000^{2i / \dmodel})
\end{align*}
where $pos$ is the sentence position and $i$ is the dimension. Using this approach, sentences longer than those seen during training can still have positional information added. We will be referring to this method as $PE$.


\subsubsection{Learned Position Encoding}

In a similar method, learned vectors of the same dimensionality, that are also unique to each position can be added to the input embeddings instead of sinusoidal position encodings\cite{gehring2017convolutional}. There are two downsides to this approach. First, this method cannot handle sentences that are longer than the ones in the training set as no vectors are trained for those positions. Second, the further position will likely not get trained as well if the training dataset has more short sentences than longer ones. \citet{Vaswani2017AttentionIA} also reported that these perform identically to the positional encoding approach.

\subsubsection{Relative Position Representations}
Relative Position Representations ($RPR$) was introduced by \cite{Shaw2018SelfAttentionWR} as a replacement of positional encodings for the Transformer. Using this approach, the Transformer was able to perform even better for NMT. Out of the three discussed, we have found this approach to work best and we will be referring to this method as $RPR$ throughout the paper.



For this method, the self-attention mechanism is modified to explicitly learn the relative positional information between every two sequence positions. As a result, the input sequence is modeled as a labeled, directed, fully-connected graph, where the labels represent positional information. A tunable parameter $k$ is also introduced that limits the maximum distance considered between two sequence positions. \cite{Shaw2018SelfAttentionWR} hypothesized that this will allow the model to generalize to longer sequences at test time.

\begin{figure*}[ht!]
\begin{center}
    \centering
    \includegraphics[width=1\textwidth]{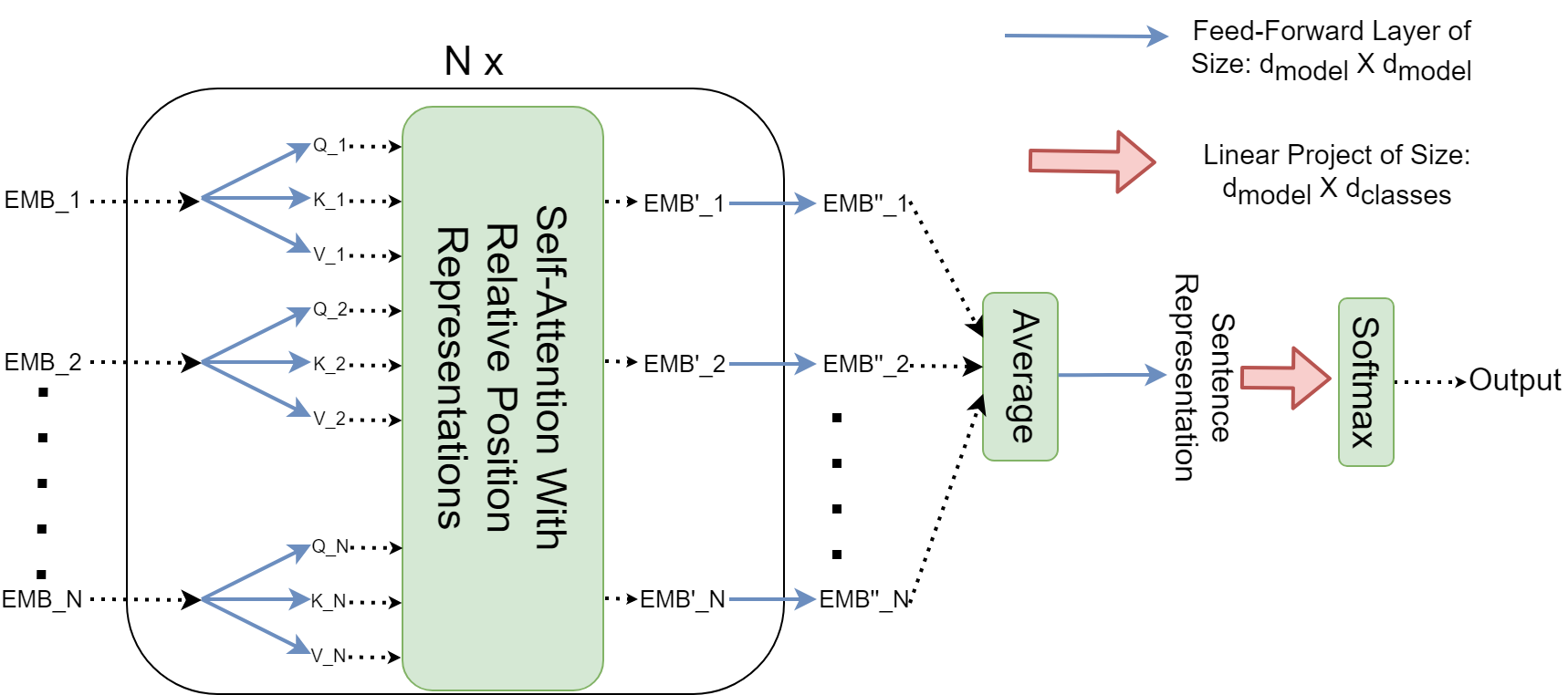}
    \caption{SSAN Model Architecture}
    \label{fig:SSAN}
\end{center}
\end{figure*}

\section{Proposed Architectures} \label{proposeMethods}

In this work we propose a simple self-attention (SSAN) model and test it in 1 as well as 2 layer stacked configurations. We designed the SSAN architecture to not have any extra components in order to compare specifically the self-attention component to the recurrence and convolution components of LSTM and CNN models. Our goal is to test the effectiveness of the main building blocks. We compare directly the results of two proposed architectures, \textit{1-Layer-SSAN} and \textit{2-Layer-SSAN}, to the LSTM, BiLSTM, and CNN architectures from \cite{SOTA_SA_2017}.

SSAN performs self-attention only once, which is identical to 1-head multi-head attention. SSAN takes in input word embeddings and applies 3 feed-forward layers to obtain Q,K,V representations on which self-attention is performed. The output of the self-attention layer is passed through another feed-forward layer. This process is done twice for \textit{2-Layer-SSAN}, using the output of first layer as input for the second. The output of the last self-attention layer is averaged and a feed-forward layer is then applied to create a sentence representation vector of fixed dimension $\dmodel$. Finally, the sentence representation vector is passed through an output softmax layer that has an output dimension of $\dclasses$. Dropout \cite{srivastava2014dropout} is applied on input word embeddings, output of self-attention layers, on the sentence representation vector. The architecture is visualized in Figure \ref{fig:SSAN}. All feed-forward layers use ReLU \cite{nair2010rectified} activation functions. For relative positional representation, we set the parameter $k$=10, which is the maximum relative position considered for each input sequence position.

Finally, we also show results for other, more complex, self-attention architectures that are based on the Transformer. We take a 2 layer Transformer encoder as described by \cite{Vaswani2017AttentionIA}, then just like for SSAN, average the output of the second layer to create a sentence representation and apply a feed-forward layer followed by an output softmax layer. Dropout is applied as described in \cite{Vaswani2017AttentionIA} as well as on the sentence representation vector.




\section{Experiments}
To reduce implementation deviations from previous work, we use the codebase from \cite{SOTA_SA_2017} and only replace the model and training process. We re-use the code for batch pre-processing and batch construction for all datasets, accuracy evaluation, as well as use the same word embeddings\footnote{\url{https://github.com/jbarnesspain/sota_sentiment}}. All neural network models use cross-entropy for the training loss.

All experiments and benchmarks were run using a single GTX 1080 Ti with an i7 5820k @ 3.3Ghz and 32Gb of RAM. For model implementations: LSTM, BiLSTM, and CNN baselines are implemented in Keras\_2.0.8 \cite{chollet2015keras} with Tensorflow\_1.7 backend using cuDNN\_5.1.5 and CUDA\_9.1. All self-attention models are implemented in Tensorflow\_1.7 and use the same CUDA libraries.



\newcommand\Tstrut{\rule{0pt}{5ex}}       
\newcommand{\smallsep}{\cmidrule(r){1-1}\cmidrule(r){2-2}\cmidrule(lr){3-3}\cmidrule(lr){4-4}\cmidrule(lr){5-5}\cmidrule(lr){6-6}\cmidrule(lr){7-7}\cmidrule(lr){8-8}\cmidrule(lr){9-9}\cmidrule(lr){10-10}}
\begin{table*}[ht]
\begin{center}
\begin{tabular}{l|rrr|ccc|ccc}
\hline &

    \MyHead{0.8cm}{Train} & \MyHead{0.3cm}{Dev.} & \MyHead{0.5cm}{Test}       & \MyHead{1.1cm}{\# of Classes}     &   \MyHead{1.3cm}{Average Sent. Length}   & \MyHead{1.1cm}{Max Sent. Length} &   \MyHead{1.3cm}{Vocab. Size} &   \MyHead{1.5cm}{Wiki Emb. Coverage} &   \MyHead{1.5cm}{300D Emb. Coverage}  \Tstrut\\
\smallsep
 \textit{SST-fine}   & 8,544 & 1,101 & 2,210  &  5 &  19.53 &  57 &   19,500 &  94.4\%  &  89.0\%  \\
 \textit{SST-binary} & 6,920 & 872   & 1,821  &  2 &  19.67 &  57 &   17,539 &  95.0\%  &  89.6\%  \\
 \textit{OpeNER}     & 2,780 & 186   & 743    &  4 &  4.28  &  23 &    2,447 &  94.2\%  &  99.3\%  \\
 \textit{SenTube-A}  & 3,381 & 225   & 903    &  2 &  28.54 & 127 &   18,569 &  75.6\%  &  74.5\%  \\
 \textit{SenTube-T}  & 4,997 & 333   & 1,334  &  2 &  28.73 & 121 &   20,276 &  70.4\%  &  76.0\%  \\
 \textit{SemEval}    & 6,021 & 890   & 2,376  &  3 &  22.40 & 40  &   21,163 &  77.1\%  &  99.8\%  \\
\hline
\end{tabular}
    \caption{Modified Table 2 from \cite{SOTA_SA_2017}. Dataset statistics, embedding coverage of dataset vocabularies, as well as splits for Train, Dev (Development), and Test sets. The 'Wiki' embeddings are the 50, 100, 200, and 600 dimension used for experiments.}
\label{table:stats}
\end{center}
\end{table*}

\subsection{Datasets}

In order to determine if certain neural network building blocks are superior, we test on six datasets from \cite{SOTA_SA_2017} with different properties. The summary for dataset properties is in Table \ref{table:stats}.

The Stanford Sentiment Treebank (\textit{SST-fine}) \cite{socher2013recursive} deals with movie reviews, containing five classes [very-negative, negative, neutral, positive, very-positive]. (\textit{SST-binary}) is constructed from the same data, except the neutral class sentences are removed, all negative classes are grouped, and all positive classes are grouped. The datasets are pre-processed to only contain sentence-level labels, and none of the models reported in this work utilize the phrase-level labels that are also provided.

The \textit{OpeNER} dataset \cite{Agerri2013} is a dataset of hotel reviews with four sentiment classes: very negative, negative, positive, and very positive. This is the smallest dataset with the lowest average sentence length.

The SenTube datasets \cite{Uryupina2014_SenTube} consist of YouTube comments with two sentiment classes: positive and negative. These datasets contain the longest average sentence length as well as the longest maximum sentence length of all the datasets.

The SemEval Twitter dataset (\textit{SemEval}) \cite{Nakov_semeval-2013task} consists of tweets with three classes: positive, negative, and neutral.


\begin{table*}[t!]
\definecolor{blue}{cmyk}{0.2,0,0,0.1}
\newcommand{\hl}[1]{{\setlength{\fboxsep}{1pt}\colorbox{blue}{#1}}}
\newcommand{\best}[1]{\textbf{\setlength{\fboxsep}{1pt}\fbox{#1}}}

\newcommand{\sepb}{\cmidrule{1-1}\cmidrule(r){2-3}\cmidrule(lr){4-4}\cmidrule(lr){5-5}\cmidrule(lr){6-6}\cmidrule(lr){7-7}\cmidrule(lr){8-8}\cmidrule(lr){9-9}\cmidrule(lr){10-10}}

\newcommand{\sep}{\cmidrule(r){2-3}\cmidrule(lr){4-4}\cmidrule(lr){5-5}\cmidrule(lr){6-6}\cmidrule(lr){7-7}\cmidrule(lr){8-8}\cmidrule(lr){9-9}\cmidrule(lr){10-10}}
  \centering
  \renewcommand*{\arraystretch}{0.8}
  \setlength\tabcolsep{2.1mm}
  \newcolumntype{P}{>{\centering\arraybackslash}p{5mm}}
\scalefont{0.85}
  \begin{tabular}{cclllllllll}
    \toprule
    & \rrt{Model} & \rrt{Dim.}&\rrt{SST-fine} & \rrt{SST-binary} & \rrt{OpeNER} & \rrt{SenTube-A} & \rrt{SenTube-T} & \rrt{SemEval} & \rrt{Macro-Avg.} \\
    \sepb
    \multirow{20}{*}{\rt{Baselines}} & 
    
    \multicolumn{1}{c}{\bow} & & 40.3 & 80.7 & 77.1 & 60.6 & 66.0 & 65.5 & 65.0\\
    \sep
    
    & \multirow{1}{*}{\averaging}
    & 300 & 41.6 & 80.3 & 76.3& 61.5& 64.3& 63.6& 64.6 \\
    \sep

  &   \multirow{5}{*}{{\lstm}} & 50 & 43.3 \sd{(1.0)} & 80.5 \sd{(0.4)}& 81.1 \sd{(0.4)}& 58.9 \sd{(0.8)}& 63.4 \sd{(3.1)}& 63.9 \sd{(1.7)}& 65.2 \sd{(1.2)}  \\ 
  &                         & 100 & 44.1 \sd{(0.8)} & 79.5 \sd{(0.6)}& 82.4 \sd{(0.5)}& 58.9 \sd{(1.1)}& 63.1 \sd{(0.4)}& 67.3 \sd{(1.1)}& 65.9 \sd{(0.7)}  \\
  &                         & 200 & 44.1 \sd{(1.6)} & 80.9 \sd{(0.6)}& 82.0 \sd{(0.6)}& 58.6 \sd{(0.6)}& 65.2 \sd{(1.6)}& 66.8 \sd{(1.3)}& 66.3 \sd{(1.1)}  \\
  &                         & 300 & 45.3 \sd{(1.9)} & 81.7 \sd{(0.7)}& 82.3 \sd{(0.6)}& 57.4 \sd{(1.3)}& 63.6 \sd{(0.7)}& 67.6 \sd{(0.6)}& 66.3 \sd{(1.0)}  \\
  & 							& 600 & 44.5 \sd{(1.4)} & 83.1 \sd{(0.9)}& 81.2 \sd{(0.8)}& 57.4 \sd{(1.1)}& 65.7 \sd{(1.2)}& 67.5 \sd{(0.7)}& 66.5 \sd{(1.0)}  \\
  \sep

  &   \multirow{5}{*}{{\bilstm}} & 50 & 43.6 \sd{(1.2)} & 82.9 \sd{(0.7)}& 79.2 \sd{(0.8)}& 59.5 \sd{(1.1)}& 65.6 \sd{(1.2)}& 64.3 \sd{(1.2)}& 65.9 \sd{(1.0)}  \\ 
  &                         & 100 & 43.8 \sd{(1.1)} & 79.8 \sd{(1.0)}& 82.4 \sd{(0.6)}& 58.6 \sd{(0.8)}& 66.4 \sd{(1.4)}&65.2 \sd{(0.6)}& 66.0 \sd{(0.9)}  \\
  &                         & 200 & 44.0 \sd{(0.9)} & 80.1 \sd{(0.6)}& 81.7 \sd{(0.5)}& 58.9 \sd{(0.3)}& 63.3 \sd{(1.0)}& 66.4 \sd{(0.3)}& 65.7 \sd{(0.6)}  \\
  &                         & 300 & 45.6 \sd{(1.6)} & 82.6 \sd{(0.7)}& 82.5 \sd{(0.6)}& 59.3 \sd{(1.0)}& 66.2 \sd{(1.5)}& 65.1 \sd{(0.9)}& 66.9 \sd{(1.1)}  \\
  &							 & 600 & 43.2 \sd{(1.1)} & 83.0 \sd{(0.4)}& 81.5 \sd{(0.5)}& 59.2 \sd{(1.6)}& 66.4 \sd{(1.1)}& 68.5 \sd{(0.7)}& 66.9 \sd{(0.9)}  \\
  \sep

  &    \multirow{5}{*}{{\cnn}} & 50 & 39.9 \sd{(0.7)} & 81.7 \sd{(0.3)}& 80.0 \sd{(0.9)}& 55.2 \sd{(0.7)}& 57.4 \sd{(3.1)}& 65.7 \sd{(1.0)}& 63.3 \sd{(1.1)}  \\ 
  &                         & 100 & 40.1 \sd{(1.0)} & 81.6 \sd{(0.5)}& 79.5 \sd{(0.9)}& 56.0 \sd{(2.2)}& 61.5 \sd{(1.1)}& 64.2 \sd{(0.8)}& 63.8 \sd{(1.1)}  \\
  &                         & 200 & 39.1 \sd{(1.1)} & 80.7 \sd{(0.4)}& 79.8 \sd{(0.7)}& 56.3 \sd{(1.8)}& 64.1 \sd{(1.1)}& 65.3 \sd{(0.8)}& 64.2 \sd{(1.0)}  \\
  &                         & 300 & 39.8 \sd{(0.7)} & 81.3 \sd{(1.1)}& 80.3 \sd{(0.9)}& 57.3 \sd{(0.5)}& 62.1 \sd{(1.0)}& 63.5 \sd{(1.3)}& 64.0 \sd{(0.9)}  \\
  &							 & 600 & 40.7 \sd{(2.6)} & 82.7 \sd{(1.2)}& 79.2 \sd{(1.4)}& 56.6 \sd{(0.6)}& 61.3 \sd{(2.0)}& 65.9 \sd{(1.8)}& 64.4 \sd{(1.5)}  \\
  \sepb


  \multirow{25}{*}{\rt{Self-Attention Models}} &    
  
  \multirow{5}{*}{{\SanOne}} 
  &                         50 & 42.8 \sd{(0.8)} & 79.6 \sd{(0.3)}& 78.6 \sd{(0.5)}& \best{64.1} \sd{(0.4)}& 67.0 \sd{(1.0)}& 67.1 \sd{(0.5)}& 66.5 \sd{(0.6)}  \\
  &                         & 100 & 44.6 \sd{(0.3)} & 82.3 \sd{(0.3)}& 81.6 \sd{(0.5)}& 61.6 \sd{(1.3)}& 68.6 \sd{(0.6)}& 68.6 \sd{(0.5)}& 67.9 \sd{(0.6)}  \\
  &                         & 200 & 45.4 \sd{(0.4)} & 83.1 \sd{(0.5)}& 82.3 \sd{(0.4)}& 62.2 \sd{(0.6)}& 68.4 \sd{(0.8)}& 70.5 \sd{(0.4)}& 68.6 \sd{(0.5)}  \\
  &                         & 300 & \best{48.1} \sd{(0.4)} & \best{84.2} \sd{(0.4)}& 83.8 \sd{(0.2)}& 62.5 \sd{(0.3)}& 68.4 \sd{(0.8)}& \best{72.2} \sd{(0.8)}& \best{69.9} \sd{(0.5)}  \\ 
  &							 & 600 & 47.7 \sd{(0.7)} & 83.6 \sd{(0.4)}& 83.1 \sd{(0.4)}& 62.0  \sd{(0.4)}&  \best{68.8} \sd{(0.7)}& 70.5 \sd{(0.8)}& 69.2 \sd{(0.5)}  \\
  \sep

  &    \multirow{5}{*}{\SanTwo} & 50 & 43.2 \sd{(0.9)} & 79.8 \sd{(0.2)}& 79.2 \sd{(0.6)}& 63.0 \sd{(1.3)}& 66.6 \sd{(0.5)}& 67.5 \sd{(0.7)}& 66.5 \sd{(0.7)}  \\ 
  &                         & 100 & 45.0 \sd{(0.4)} & 81.6 \sd{(0.9)}& 81.1 \sd{(0.4)}& 63.3 \sd{(0.7)}& 67.7 \sd{(0.5)}& 68.7 \sd{(0.4)}& 67.9 \sd{(0.5)}  \\
  &                         & 200 & 46.5 \sd{(0.7)} & 82.8 \sd{(0.5)}& 82.3 \sd{(0.6)}& 61.9 \sd{(1.2)}& 68.0 \sd{(0.8)}& 69.6 \sd{(0.8)}& 68.5 \sd{(0.8)}  \\
  &                         & 300 & \best{48.1} \sd{(0.8)} & 83.8 \sd{(0.9)}& 83.3 \sd{(0.9)}& 62.1 \sd{(0.8)}& 67.8 \sd{(1.0)}& 70.7 \sd{(0.5)}& 69.3 \sd{(0.8)}  \\ 
  &							 & 600 & 47.6 \sd{(0.5)} & 83.7 \sd{(0.4)}& 82.9 \sd{(0.5)}& 60.7 \sd{(1.4)}& 68.2 \sd{(0.7)}& 70.3 \sd{(0.3)}& 68.9 \sd{(0.6)}  \\
  \sep

  &    \multirow{3}{*}{\Transf} & \multirow{3}{*}{300} & \multirow{3}{*}{47.3 \sd{(0.4)}} & \multirow{3}{*}{83.8 \sd{(0.4)}}& \multirow{3}{*}{\best{84.2} \sd{(0.5)}}& \multirow{3}{*}{62.0 \sd{(1.4)}}& \multirow{3}{*}{68.2 \sd{(1.6)}}& \multirow{3}{*}{72.0 \sd{(0.5)}}& \multirow{3}{*}{69.6 \sd{(0.8)}}  \\ \\ \\

 
 \sep
 

 &    \multirow{3}{*}{\TransfWithPosEnc} & \multirow{3}{*}{300} & \multirow{3}{*}{45.0 \sd{(0.7)}} & \multirow{3}{*}{82.0 \sd{(0.6)}}& \multirow{3}{*}{83.3 \sd{(0.7)}}& \multirow{3}{*}{62.3 \sd{(2.4)}}& \multirow{3}{*}{66.9 \sd{(0.8)}}& \multirow{3}{*}{68.4 \sd{(0.8)}}& \multirow{3}{*}{68.0 \sd{(1.0)}}  \\ \\ \\
 
 \sep
 
 &    \multirow{3}{*}{\SanOneNoPos} & \multirow{3}{*}{300} & \multirow{3}{*}{47.2 \sd{(0.5)}} & \multirow{3}{*}{83.9 \sd{(0.7)}}& \multirow{3}{*}{83.6 \sd{(0.6)}}& \multirow{3}{*}{62.1 \sd{(2.5)}}& \multirow{3}{*}{68.7 \sd{(1.0)}}& \multirow{3}{*}{70.2 \sd{(1.2)}}& \multirow{3}{*}{69.3 \sd{(1.1)}}  \\ \\ \\
 
 \sep
 
  &    \multirow{2}{*}{\SanOneWithPosEnc} & \multirow{2}{*}{300} & \multirow{2}{*}{45.0 \sd{(0.3)}} & \multirow{2}{*}{82.9 \sd{(0.2)}}& \multirow{2}{*}{80.7 \sd{(0.6)}}& \multirow{2}{*}{62.6 \sd{(2.3)}}& \multirow{2}{*}{67.8 \sd{(0.4)}}& \multirow{2}{*}{69.1 \sd{(0.3)}}& \multirow{2}{*}{68.0 \sd{(0.7)}}  \\ \\

  \bottomrule
  \end{tabular}
  \caption{Modified Table 3 from \cite{SOTA_SA_2017}. Test accuracy averages and standard deviations (in brackets) of 5 runs. The baseline results are taken from \cite{SOTA_SA_2017}; the self-attention models results are ours. Best model for each dataset is given in \best{bold}.} \label{table:results}

\end{table*}


\newcommand\TstrutTwo{\rule{0pt}{4ex}} 
\newcommand{\smallseptwo}{\cmidrule(r){1-1}\cmidrule(r){2-2}\cmidrule(lr){3-3}\cmidrule(lr){4-4}\cmidrule(lr){5-5}}
\begin{table*}[t!]
\begin{center}
\begin{tabular}{lcccc}
\hline
    \multicolumn{1}{c}{Model} & \multicolumn{1}{c}{\# of Parameters} & \multicolumn{1}{c}{GPU VRAM Usage (MB)}       & \MyHead{1.8cm}{Training Time (s)}     &   \MyHead{2.5cm}{Inference Time (s)}   \TstrutTwo\\
\smallseptwo
 \textit{LSTM}                  & 722,705       & 419MB     & 235.9s      &  7.6s \\
 \textit{BiLSTM}                & 1,445,405     & 547MB     & 416.0s      &  12.7s \\
 \textit{CNN}                   & 83,714        & 986MB     & 21.1s       &  0.85s \\
 \textit{1-Layer SSAN + RPR}    & 465,600       & 381MB     & 64.6s       &  8.9s \\
 \textit{1-Layer SSAN + PE}     & 453,000       & 381MB     & 58.1s       &  8.5s \\
 \textit{2-Layer SSAN + RPR}    & 839,400       & 509MB     & 70.3s       &  9.3s \\
 \textit{Transformer + RPR}     & 1,177,920     & 510MB     & 78.2s       &  9.7s \\
\hline
\end{tabular}
    \caption{Neural networks architecture characteristics. A comparison of number of learnable parameters, GPU VRAM usage (in megabytes) during training, as well as training and inference times (in seconds).} \label{table:characteristics}
\label{table:perf}
\end{center}
\end{table*}


\subsection{Embeddings}

We use the exact same word embeddings as \cite{SOTA_SA_2017}. They trained the 50, 100, 200, and 600-dimensional word embeddings using the word2vec algorithm described in \cite{mikolov2013efficient} on a 2016 Wikipedia dump. In order to compare to previous work, they also used the publicly available Google 300-dimensional word2vec embeddings, which are trained on a part of Google News dataset\footnote{\url{https://code.google.com/archive/p/word2vec/}}. For all models, out-of-vocabulary words are initialized randomly from the uniform distribution on the interval [-0.25 , 0.25].





\subsection{Baselines}


We take 5 classifiers from \cite{SOTA_SA_2017} and use their published results as baselines. Two of the methods are based on logistic regression, \bow and \averaging, and 3 are neural network based, \lstm, \bilstm, and \cnn.

The (\bow) baseline is a L2-regularized logistic regression trained on bag-of-words representation. Each word is represented by a one-hot vectors of size $n = |V|$, where $|V|$ is the vocabulary size.

The (\averaging) baseline is also a L2-regularized logistic regression classifier except trained on the average of the 300-dimension word embeddings for each sentence. 

The \lstm baseline, input word embeddings are passed into an LSTM layer. Then a 50-dimensional feed-forward layer with ReLU activations is applied, followed by a softmax layer that produces that model classification outputs. Dropout \cite{srivastava2014dropout} is applied to the input word embeddings for regularization.

The \bilstm baseline is the same as \lstm, except that a second LSTM layer is used to process the input word embeddings in the reverse order. The outputs of the two LSTM layers are concatenated and passed a feed-forward layer, following by the output softmax layer. Dropout is applied identically as in \lstm. This modification improves the networks ability to capture long-range dependencies. 

The final baseline is a simple \cnn network. The input sequence of $n$ embeddings is reshaped to an $n\times R$ dimensional matrix $M$, where $R$ is the dimensionality of the embeddings. Convolutions with filter size of [2,3,4] are applied to $M$, following by a pooling layer of length 2. As for \lstm networks, a feed-forward layer is applied followed by an output softmax layer. Here, dropout is applied to input embeddings as well as after the convolution layers.

The \lstm, \bilstm, and \cnn baselines are trained using ADAM \cite{Kingma2014AdamAM} with cross-entropy loss and mini-batches of size 32. Hidden layer dimension, dropout amount, and the number of training epochs are tuned on the validation set for each (model, input embedding, dataset) combination.

\subsection{Self-Attention Architectures}

We use \textit{1-Layer SSAN + RPR} and \textit{2-Layer SSAN + RPR} to compare the self-attention mechanism to the recurrence and convolution mechanisms in \textit{LSTM}, \textit{BiLSTM}, and \textit{CNN} models. We compare these models using all word embeddings sizes. 

Next, we explore the performance of a modified \textit{Transformer Encoder} described in \ref{proposeMethods}. We do this to determine if a more complex architecture that utilized multi-head attention is more beneficial.

Finally, we compare the performance of using positional encodings (\textit{+PE}) and relative positional representations (\textit{+RPR}) for the \textit{Transformer Encoder} and \textit{1-Layer-SSAN} architectures. We also test \textit{1-Layer SSAN} without using any positional information techniques.

For the self-attention networks, we simplify the training process to only tune one parameter and apply the same process to all models. Only the learning rate is tuned for every (model, input embedding) pair. We fix the number of batches to train for to 100,000 and pick the model with highest validation accuracy. Each batch is constructed by randomly sampling the training set. Model dimensionality $\dmodel$ is fixed to being the same as the input word embeddings. Learning rate is tuned based on the size of $\dmodel$. For $\dmodel$ dimensions [50, 100, 200, 300, 600] we use learning rates of [0.15, 0.125, 0.1, 0.1, 0.05] respectively, because the larger $\dmodel$ models tend to over-fit faster. Dropout of 0.7 is applied to all models, and the ADADELTA \cite{zeiler2012adadelta} optimizer is used with cross-entropy loss.






\section{Analysis}

Table \ref{table:results} contains the summary of all the experimental results. For all neural network models we report mean test accuracy of five runs as well as the standard deviations. Macro-Avg results are the average accuracy of a model across all datasets. We focus our discussion on the Macro-Avg column as it demonstrates the models general performance for sentiment analysis.


Our results show general better performance for self-attention networks in comparison to \lstm, \bilstm and \cnn models. Using the same word embedding, all of the self-attention models receive higher Macro-Avg accuracy than all baseline models. \textit{1-Layer-SSAN+RPR} models generally perform the best for all (input embeddings, dataset) combinations, and getting top scores for five out of six datasets. \textit{Transformer Encoder+RPR} also performs comparatively well across all datasets, and achieves top accuracy for the \textit{OpeNER} dataset.

Using \textit{2-Layer-SSAN+RPR} does not yield better performance results compared to \textit{1-Layer-SSAN+RPR}. We believe that one self-attention layer is sufficient as the datasets that we have tested on were relatively small. This is reinforced by the results we see from \textit{Transformer Encoder + RPR} since it achieves similar accuracy as \textit{2-Layer-SSAN+RPR} and \textit{1-Layer-SSAN+RPR} while having greater architectural complexity and more trainable parameters, see Table \ref{table:perf}.

Using relative positional representations for \textit{1-Layer-SSAN+RPR} increases the Macro-Avg accuracy by 2.8\% compared to using positional encodings for \textit{1-Layer-SSAN+PE}, and by 0.9\% compared to using no positional information at all (\textit{1-Layer-SSAN}). Interestingly enough, we observe that using no positional information performs better than using positional encodings. This could be attributed once again to small dataset size, as \cite{Vaswani2017AttentionIA} successfully used positional encodings for larger MT datasets.

Another observation is that SenTube dataset trials achieve a low accuracy despite having binary classes. This is unexpected as generally with a low number of classes it is easier to train on the dataset and achieve higher accuracy. We suspect that this is because SenTube contains longer sentences and very low word embedding coverage. Despite this, SSANs perform relatively well on the SenTube-A dataset, which suggests that they are superior at capturing long-range dependencies compared to other models.

Smaller $\dmodel$ \textit{SSAN} models perform worse for lower dimension input embeddings on \textit{SST-fine}, \textit{SST-binary} and \textit{OpeNER} datasets while still performing well on \textit{SenTube} and \textit{SemEval}. This is caused by the limitations of our training process where we forced the network $\dmodel$ to be same size as the input word embeddings and use the same learning rate for all datasets. We found that working with smaller dimensions of $\dmodel$ the learning rate needed to be tuned individually for some datasets. For example, using a learning of 0.15 for 50D models would work well for \textit{SenTube} and \textit{SemEval}, but would under-fit for \textit{SST-fine}, \textit{SST-binary} and \textit{OpeNER} datasets. We decided to not modify the training process for the smaller input embeddings in order to keep our training process simplified.

\subsection{Model Characteristics}
Here we compare training and test efficiency, memory consumption and number of trainable parameters for every model. For all models, we use the \textit{SST-fine} dataset, hidden dimension size of 300, Google 300D embeddings, batch sizes of 32 for both training and inference, and the ADAM optimizer \cite{Kingma2014AdamAM}. The \textit{Training Time} test is the average time it takes every model to train on 10 epochs of the SST-fine train set (~2670 batches of size 32). The \textit{Inference Time} test is the average time it takes a model to produce predictions for the validation set 10 times (~344 batches of size 32). Table \ref{table:perf} contains the summary of model characteristics. The GPU VRAM usage is the amount of GPU video memory that is used during training. 

\cnn has the lowest number of parameters but consumes the most GPU memory. It also has the shortest training and inference time, which we attributed to the low number of parameters.

Using relative position representations compared to positional encoding for \textit{1-Layer-SSAN} increases the number of trainable parameters by only 2.7\%, training time by 11.2\%, and inference time by 4.7\%. These findings are similar to what \cite{Shaw2018SelfAttentionWR} reported.

\textit{BiLSTM} has double the number of parameters as well as near double training and inference times compared to \textit{LSTM}. This is reasonable due to the nature of the architecture being two \textit{LSTM} layers. Much like \textit{BiLSTM}, going from \textit{1-Layer-SSAN} to \textit{2-Layer-SSAN} doubles the number of trainable parameters. However, the training and inference times only increase by 20.1\% and 9.4\% respectively. This demonstrates the efficiency of the self-attention mechanism due to it utilizing only matrix multiply operations, for which GPUs are highly-optimized.

We also observe that self-attention models are faster to train than \textit{LSTM} by about 3.4 times, and 5.9 times for \textit{BiLSTM}. However, inference times are slower than \textit{LSTM} by 15.5\% and faster than \textit{BiLSTM} by 41\%.

\section{Conclusion}

In this paper we focused on demonstrating that self-attention networks achieve better accuracy than previous state-of-the-art techniques on six datasets. 
In our experiments, multiple \textit{SSAN} networks performed better than \textit{CNN} and \textit{RNN} architectures; Self-attention architecture resulted in higher accuracy than \textit{LSTMs} while having 35\% fewer parameters and shorter training time by a factor of 3.5. Additionally, we showed that SSANs achieved higher accuracy on the \textit{SenTube} datasets, which suggests they are also better at capturing long-term dependencies than \textit{RNNs} and \textit{CNNs}.
Finally, we reported that using relative positional representation is superior to both using positional encodings, as well as not incorporating any positional information at all. Using relative positional representations for self-attention architectures resulted in higher accuracy with negligible impact on model training and inference efficiency.

For future work, we plan to extend the \textit{SSAN} networks proposed to achieve state-of-the-art results on the complete \textit{SST} dataset. We are also interested to see the behaviour of the models explored in this work on much larger datasets, we hypothesize that stacked multi-head self-attention architectures will perform significantly better than \textit{RNN} and \textit{CNN} counterparts, all while remaining more efficient at training and inference.


\section*{Acknowledgments}
We thank the anonymous reviewers for their insightful suggestions.

\bibliographystyle{plainnat}
\bibliography{references}
\end{document}